# Wind speed forecast using random forest learning method

G. V. Drisya[a], Valsaraj P.[a], K. Asokan[b], K. Satheesh Kumar[a,*]

[a] Department of Futures Studies, University of Kerala, Kariavattom, Kerala, India - 695 581
[b] Department of Mathematics, College of Engineering, Thiruvananthapuram, Kerala, India - 695 016
Email: drisyavictoria@gmail.com , valsaraj@hotmail.com, asokank@cet.ac.in , *
kskumar@keralauniversity.ac.in

*Abstract—* Wind speed forecasting models and their application to wind farm operations are attaining remarkable attention in the literature because of its benefits as a clean energy source. In this paper, we suggested the time series machine learning approach called random forest regression for predicting wind speed variations. The computed values of mutual information and auto-correlation shows that wind speed values depend on the past data up to 12 hours. The random forest model was trained using ensemble from two weeks data with previous 12 hours values as input for every value. The computed root mean square error shows that model trained with two weeks data can be employed to make reliable short-term predictions up to three years ahead.

## I. INTRODUCTION

Surface wind plays a crucial role in the economic development of a country as it is widely recognised as a clean, inexpensive, inextinguishable alternate source of energy. All over the world, generation of electricity from the wind has steadily been increased over the last few years, and it is estimated that, at the end of 2016 total installed wind capacity is 487 GW, and in a moderate scenario by 2020 it will reach 792 GW [1]. In a local scenario, the total nationwide installed wind power capacity by the end of 2015 were 47 GW providing almost 67% of renewable energy connected to grid. India is also expecting 125 GW of installed capacity by 2030 with the aim of 82 million tonnes of reduction in $CO_2$ emissions [2]. For an effective implementation of the strategic plans, wind energy sector confronts many challenges and proper research and development activities are needed to address these issues. The International Energy Agency Wind (IEAWind) [3], frequently lists the challenges as the numbered R&D tasks and shares among members to participate in collaborative research activities. Wind characterization is one among the list, and it addresses the research needs associated with site optimisation, operation of wind turbine and power plant and performance & output prediction. Reducing the performance uncertainties of wind power plants became more crucial, and it is important to know the rate of variation of produced power on different lead time hours. Since wind power production of a turbine is a direct function of wind speed, any improvement in existing wind speed forecasting models can contribute to wind energy industry to bring down the energy cost. Corresponding to the prediction 20 horizon, wind forecasting can be broadly classified as, very short-term (few seconds to 30 minutes), short-term (30 minutes to 6 hours), medium-term (6 hours to 1 day ahead), and long-term forecasting (1 day to 1 week) [3]. Although both physical and time series models are available for wind speed prediction, it is reported that for a very short-term to short-term forecast length, linear time series models are suitable [4]. Assuming the wind speed variations as purely a random process, statistical methods like Gaussian distribution, AR, ARMA, ARIMA, etc. were employed to explain the variations over time. For improving the prediction accuracy hybrid methods which combine different techniques such as statistical and data mining were also evaluated. Suitability of deterministic methods in predicting short-term wind speed is also investigated and reported by [5]. Machine learning techniques such as artificial neural network (ANN), support vector regression (SVR), etc. are gaining attention in recent past in the literature for wind speed prediction [6–8]. However, most such techniques employed in the literature were restricted to one step ahead prediction [9].The random forest method has employed for monitoring wind turbines [10] and forecasting wind power [11]. In this paper, we investigate the suitability of random forest models trained on a wind speed time series of short period for prediction in the subsequent years. Our results show that a random forest model trained with wind speed time series of two weeks can be utilised to make the prediction with test data almost three years ahead without significant reduction in prediction accuracy.

## II. RANDOM FOREST METHOD

Random forest is a non-parametric ensemble based learning technique used for both classification and regression problem and is first suggested by Leo Beriman [12]. It is an extended version of decision tree algorithm [13] which works on a set of rules and the possible outcomes to form a tree-like structure. For an incorrect rule adds the impurity to the subsequent nodes, a high risk of error propagation is always associated with decision trees. Random forest algorithm eliminates error diffusion property inherent in decision trees by constructing multiple decision trees. Random samples of given data set are generated and fed to several tree-based learners to form a random forest. Splitting condition for each node in a tree is based on only the randomly selected predictor attributes which lower the error rate by avoiding the correlation among the trees. In short, random forest algorithm is an extended version of an ensemble learning method called bagging in which averaging of trees is done cleverly [14]. The successful application of random forest regression algorithm has already been reported in many fields like cheminformatics, speech recognition, bioinformatics, classification and prediction in ecology, analysis of complex remote sensing datasets etc.[15–19].

Random forest regression is a non-parametric regression technique in which the functional relation between dependent and independent variables are captured from the features of the data. From a given data set, algorithm generates a forest of $N$ trees, $\{T_1(X), T_2(X), T_3(X), …, T_N(X)\}$, using a $m$ dimension vector input $X = \{x_1, x_2, …, x_m\}$. For each tree $T(x)$, the generated $N$ outcomes are represented as,

$$\hat{S} = T_1(X), \hat{S} = T_2(X), …, \hat{S}_N = T_N(X)$$

Where $\hat{S}_N$ is the output of $n^{th}$ tree. Average of all individual regression tree output is considered as the response from the random forest. In simple words random forest algorithm can be explained by a three-step procedure as follows:

1. From the given data select random samples with replacement
2. At each level, split node properly to get the best split until a maximum level of tree is obtained
3. Repeat the second step until a satisfied number of trees are generated.

Random selection with replacement is commonly known as bagging process and in random forest algorithm, the process is done both on the samples and on the attribute. Normally two third of the data will be the size of bootstrap samples, and the rest is known as out-of-bag samples. Combined effect of bootstrap and attribute bagging helps the algorithm to reduce misclassification error.

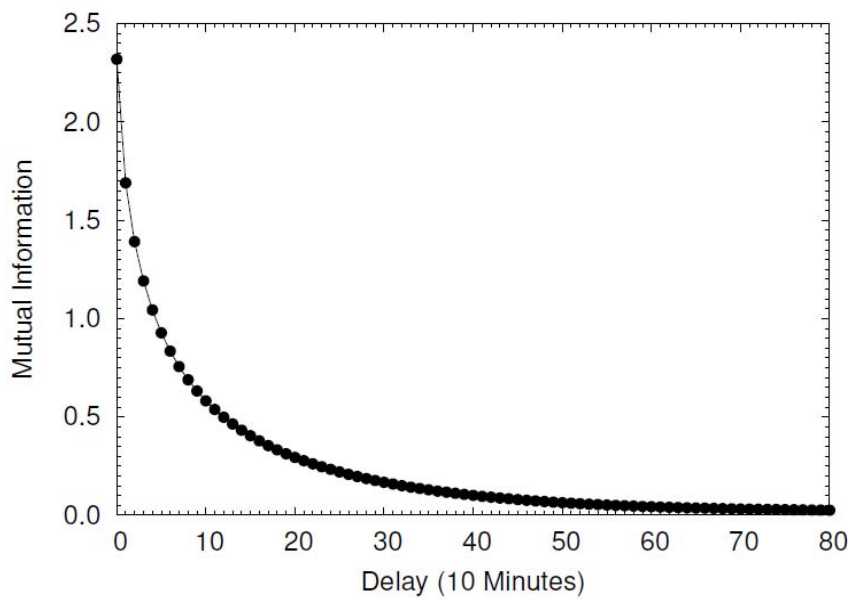

Figure 1: Mutual information as a function of delay obtained for wind speed time series

### III. RESULTS AND DISCUSSIONS

The main objective of this paper is to investigate the performance of random forest regression model to forecast wind speed measurements. We also assess the model prediction error as time separation between train and test data increases. The data we used in this work is the measurements wind speed at 10 minutes interval at height of 80 metres at location latitude 34.98420° longitude -104.03971° from January 1, 2004 to December 31, 2016.

As a first step, since the random forest regression algorithm uses vector input for model building, a matrix of time dependent chunks of data is generated using time delay information. The dependency of a value in a time series on the previous values can be estimated by autocorrelation function. Another method of estimating the period of dependence is the calculation of mutual information between delayed time series [20]. For various delays, the average mutual information of two random variable $X$ and $Y$ is computed as

$$I(\tau) = \sum_{x \in X} \sum_{y \in Y} p(x,y) \log_2 \frac{p(x,y)}{p(x)p(y)} \quad (1)$$

Where $p(x,y)$ is the joint probability and $p(x)$ and $p(y)$ are the marginals. As can be seen from Fig. 1, mutual information computed for various delays vanishes by 72 data points (12 hours). Therefore, we consider every wind speed data point is a function of its previous 72 data points and is represented as

$$w_{i+73} = f(w_{i+1}, w_{i+2}, \ldots, w_{i+72}) \quad (2)$$

where $w_i$ for $i = 1,2,3,\ldots$ is the wind speed time series. The success of any data mining model depends on the clever sampling of training data set. Instead of having disjoint segments, we selected $(w_{i+1}, w_{i+2}, \ldots, w_{i+72})$ as the input vector with $w_{i+73}$ as the target for $i = 1,2,3,\ldots,N$. We selected two weeks data points ($N = 2016$) as training data set and subsequent 2016 data points as validation data set. The trained model was used to predict the remaining data points at an interval two weeks. The prime focus of this analysis is to evaluate the nature of prediction error as the prediction move away from the train data set. The root mean square error was calculated by

$$RMSE = \sqrt{\frac{1}{n} \sum_{i=1}^{n} (w_i^o - w_i^p)} \quad (3)$$

where $w_i^o$ and $w_i^p$ are the original and predicted wind speeds. After training and validation of the model, we obtained 10 minutes ahead prediction for the test data set. We obtained one step ahead predictions of 2016 data points at an interval of two weeks from second month onwards upto the third year. Predictions from 4032 and 150001 data points are given in Fig. 3. Note that each forecast value was obtained using previous 72 original data points. That is, we utilized only the data of first month for training and validation of the total duration three year of the given wind speed time series. The 1-step ahead (10 minutes) prediction throughout test data of 35 months shows similar accuracy as in Fig. 3. It may be noted that at peak values model tend to underestimate the speed. The support vector prediction computed by D. Liu [9] also shows a similar trend.

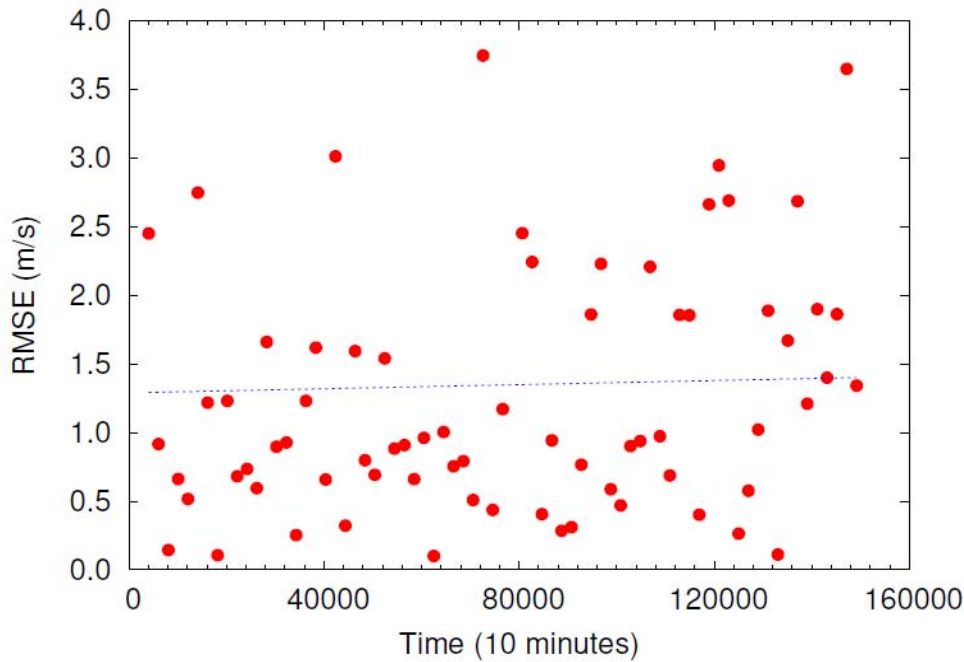

Figure 2: Plot of the root mean square error (RMSE) of one hour ahead prediction at the interval of two weeks. The dotted blue line shows the fit of the data.

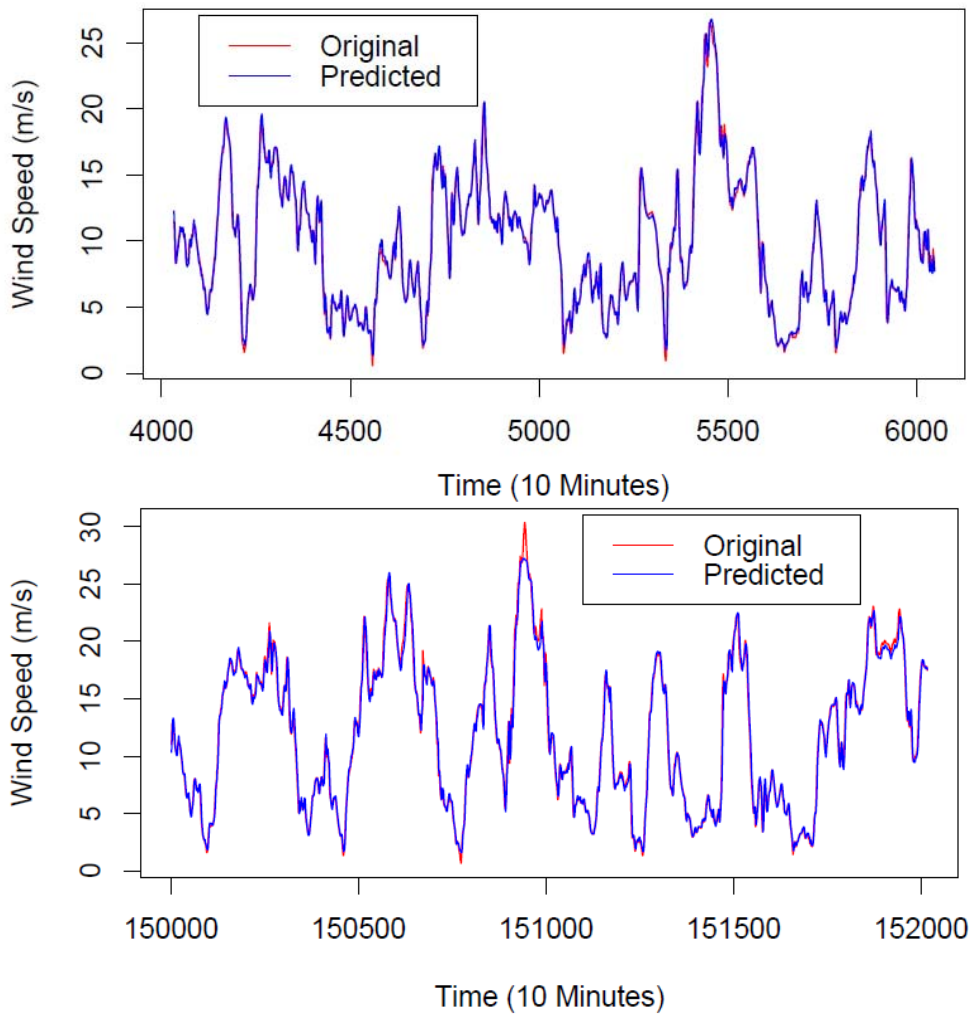

Figure 3: Comparison between the actual speed and the 10-minutes-ahead forecast by random forest algorithm.

Next, we computed one hour ahead predictions by making six one step ahead predictions by appending predicted to input vector at each step. The one–hour predictions were computed throughout the test data at an interval of two weeks. The root mean square error (RMSE) of one–hour predictions are given in Fig. 2 and the average RMSE is observed as 1.49 m/s. It may also be noted that the regression line (dotted blue line) of the RMSE shows almost negligible increase slope with time. This is interesting that a model trained with two weeks of data is capable of predicting with almost same average even in the third year. In most of the similar work reported in literature the training data is lengthier or test data is set closer to the training data set in time [9]. Sample one hour ahead predictions closer and away from train data set are given in Fig. 4. Similar plots for 2 hour ahead predictions are given in Fig. 5. The root mean square error (RMSE) two hour ahead predictions are given in Fig. 6 and the average RMSE is 2.06 m/s.

## IV. CONCLUSION

This paper presented the results of random forest regression approach for modelling and prediction of wind speed variations. The analysis shows that with the proper time delay embedding matrix random forest algorithm offers a reliable short-term prediction with two weeks training data. The model trained with two weeks data shown remarkable prediction accuracy for test data upto three years ahead of training data set in time. The root mean square error calculated for different prediction horizons at different points show not much variation in the accuracy of the model. From the results presented here, it is clear that random forest method is a suitable candidate for an effective wind speed prediction and thereby helping an efficient, cost effective wind energy management.

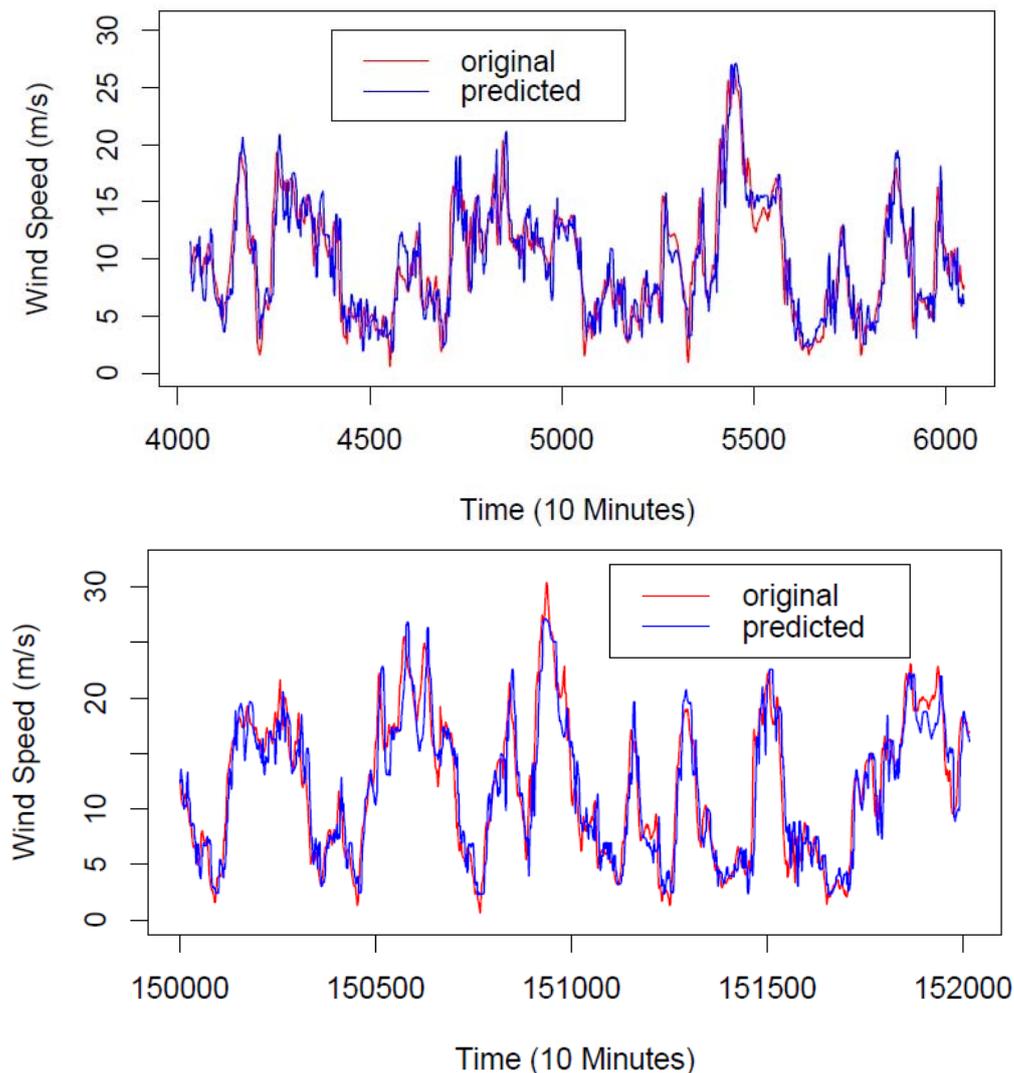

Figure 4: Comparison between the actual speed and the forecast values after one hour by random forest algorithm

ACKNOWLEDGMENT (HEADING 5)

The authors are grateful to University Grants Commission (UGC), New Delhi for their financial assistance (No. F. 42-30/2013(SR)). They wish to thank National Renewable Energy Laboratory (http://www.nrel.gov), USA for making their data available and campus computing facility of University of Kerala set up under DST-PURSE programme for providing computational facilities.

## AUTHORS PROFILE

**G. V. Drisya** is a Ph. D. student in the Department of Futures Studies at the University of Kerala, India with particular interests in chaotic time series, wind speed dynamics and Linux shell scripting. She holds a master's degree in Technology Management from University of Kerala and a bachelor's degree in Information Technology from Mahatma Gandhi University, India. Her research work focus on analysis of deterministic characteristics of wind speed dynamics with a particular attentiveness in short-term prediction.

**Valsaraj P.** is a mechanical engineer with his B. Tech. Degree and M. Tech. Degree from the University of Calicut. He has been working for the past twenty-two years as a scientist in the field of renewable energy, with Agency for Non-conventional Energy and Rural Technology (ANERT), Thiruvananthapuram, an institution established by the state government of Kerala, under its power department. Currently, he is pursuing his Ph. D. research programme in the area of wind forecasting at Department of Futures Studies, University of Kerala as a part time research fellow.

**K. Asokan** received his MSc. in Mathematics from Maharaja Sayajirao University of Baroda in 1993, M. Phil. In 1996 from University of Kerala and his Ph. D. under Faculty of Technology from Cochin University of Science and Technology, India in 2004. Currently, he is an Associate Professor in the Department of Mathematics in College of Engineering, Trivandrum, Kerala, India and his research interests include Brownian dynamics, social networks, and wind speed oscillation dynamics.

**K. Satheesh Kumar** received his MSc. in Mathematics from Annamalai University, India in 1988 and his Ph. D. under Faculty of Technology from Cochin University of Science and Technology, India in 1998. Currently, he is an Assistant Professor in the Department of Futures Studies in University of Kerala, India and his research interests include computational modelling and simulations in dynamics and rheology of suspensions and polymer solutions, time series analysis in geophysics and wind dynamics and forecasting